\documentclass[conference]{IEEEtran}
\IEEEoverridecommandlockouts
\usepackage{cite}
\usepackage{amsmath,amssymb,amsfonts}
\usepackage{algorithmic}
\usepackage{graphicx}
\usepackage{textcomp}
\usepackage{xcolor}

\usepackage{booktabs,siunitx}

\usepackage{subcaption}
\usepackage[export]{adjustbox}
\usepackage{multirow}

\DeclareMathOperator*{\argmin}{\text{arg}\min}

\usepackage{commath}
\usepackage[linesnumbered,ruled,vlined]{algorithm2e}
\SetKwInput{KwInput}{Input}                
\SetKwInput{KwOutput}{Output}              

\def\BibTeX{{\rm B\kern-.05em{\sc i\kern-.025em b}\kern-.08em
    T\kern-.1667em\lower.7ex\hbox{E}\kern-.125emX}}
\begin{document}

\title{Adverse Weather Conditions Augmentation of LiDAR Scenes with Latent Diffusion Models\\
}

\author{

\IEEEauthorblockN{1\textsuperscript{st} Andrea Matteazzi\textsuperscript{1,2}}
\IEEEauthorblockA{
matteazzi@uni-wuppertal.de}

\and 

\IEEEauthorblockN{2\textsuperscript{nd} Pascal Colling\textsuperscript{2}}
\IEEEauthorblockA{
pascal.colling@aptiv.com}

\and 

\IEEEauthorblockN{3\textsuperscript{rd} Michael Arnold\textsuperscript{2}}
\IEEEauthorblockA{
michael.arnold@aptiv.com}

\and 

\IEEEauthorblockN{4\textsuperscript{th} Dietmar Tutsch\textsuperscript{1}}
\IEEEauthorblockA{
tutsch@uni-wuppertal.de}

\and
\centerline{\textsuperscript{1}\textit{University of Wuppertal}}
\and
\centerline{\textsuperscript{2}\textit{Aptiv Services Deutschland GmbH}}

}


\maketitle

\begin{abstract}
LiDAR scenes constitute a fundamental source for several autonomous driving applications. Despite the existence of several datasets, scenes from adverse weather conditions are rarely available. This limits the robustness of downstream machine learning models, and restrains the reliability of autonomous driving systems in particular locations and seasons. Collecting feature-diverse scenes under adverse weather conditions is challenging due to seasonal limitations. Generative models are therefore essentials, especially for generating adverse weather conditions for specific driving scenarios. In our work, we propose a latent diffusion process constituted by autoencoder and latent diffusion models. Moreover, we leverage the clear condition LiDAR scenes with a postprocessing step to improve the realism of the generated adverse weather condition scenes. 
\end{abstract}


\section{Introduction}
LiDAR-based applications such as $3$D object detection must be reliable across different scenes and weather conditions. To achieve this degree of robustness, machine learning models need to be trained on a massive amount of feature-diverse data. For example, $3$D object detection models may fail on detecting heavily occluded objects or false detecting clusters of snow as objects  \cite{dreissig2023survey}. However, getting specific driving scenarios in particular adverse weather conditions can be challenging and time consuming. For example, collecting specific scenarios in heavy snowy conditions not only demands specific seasonal time, that can be rarely achieved across years, but it can also cause LiDAR sensor contamination from snow accumulation. Additionally, manually annotating ground truth can also be challenged by heavy occlusion of objects. For this reason, several datasets \cite{geiger2012we,caesar2020nuscenes} lack adverse weather conditions scenarios. On the other hand, previous research tried to work on the other way around and denoising adverse weather scenarios, as in the case of snowfall in \cite{charron2018noising}.
While this work successfully removes snow clutter noise, it still suffers from heavy occluded objects. Compared to methods based on denoising, adverse weather data augmentation methods demonstrate superior effectiveness and accuracy in terms of downstream perception performance \cite{zhang2024lidar,hahner2022lidar}.
In our work, we focus on the generation of heavy snowy conditions augmentation data. We work with the Boreas dataset \cite{burnett2023boreas}, as it contains a repeated route in both sunny and snowy conditions. However, we do not explicitly leverage the availability of close matching sunny and snowy scenes during the training, but rather for the evaluation of our method. Our method can work with diverse sunny and snowy scenes and it allows generalization capabilities of the adverse weather conditions augmentation. Our method comprises a latent diffusion process, including novel autoencoder and latent diffusion models (LDMs), aiming to recover the structure and add the adverse weather conditions to clear weather scenarios. A postprocessing step is further derived in order to recover fine-grained details of the generated adverse weather scenes, leveraging the corresponding clear weather input scenes.
Because of the complex details of heavy snowy conditions, we (plan to) quantitatively prove the effectiveness of our augmentation method with a $3$D object detection model, CenterPoint \cite{yin2021center}. We separately train the model with sunny scenes augmented with our generated snow and with only sunny scenes, and validate both the models on real snowy scenarios. The repeated route among both the weather conditions scenes allows to disentangle the generalization capabilities of the trained model across different scenarios, to the capacity of the model to learn the generated snow distribution patterns. This permits to directly correlate improvements of the model, trained on the augmented data, to the capability of our method to generate snow that resembles the real one.

In summary, our key contributions are:
\begin{itemize}
\item We propose a novel procedure to augment clear scenes with adverse weather conditions.
\item We introduce a novel autoencoder and conditioned latent diffusion model to learn distributions of adverse weather conditions.
\item We introduce a novel postprocessing method to leverage the input clear weather scenes and improve the generated adverse weather scenes with fine-grained details.
\item We (plan to) validate our components with distance-based and statistical metrics 
and we (plan to) validate our augmentation with $3$D object detection.
\end{itemize}

\section{Related Work}
\subsection{Adverse Weather Conditions}
The problem of generating adverse weather conditions for LiDAR scenes is still under explored. Previous works \cite{zhang2024lidar,zhang2023dig} tried to generate snowy conditions with methods based on CycleGAN \cite{zhu2017unpaired}. However, diffusion models \cite{ho2020denoising} demonstrate better generation results than generative adversarial networks (GANs) in image synthesis \cite{dhariwal2021diffusion}.


\section{Our Method}
\begin{figure}[h]
\centering
\includegraphics[width=1.0\linewidth]{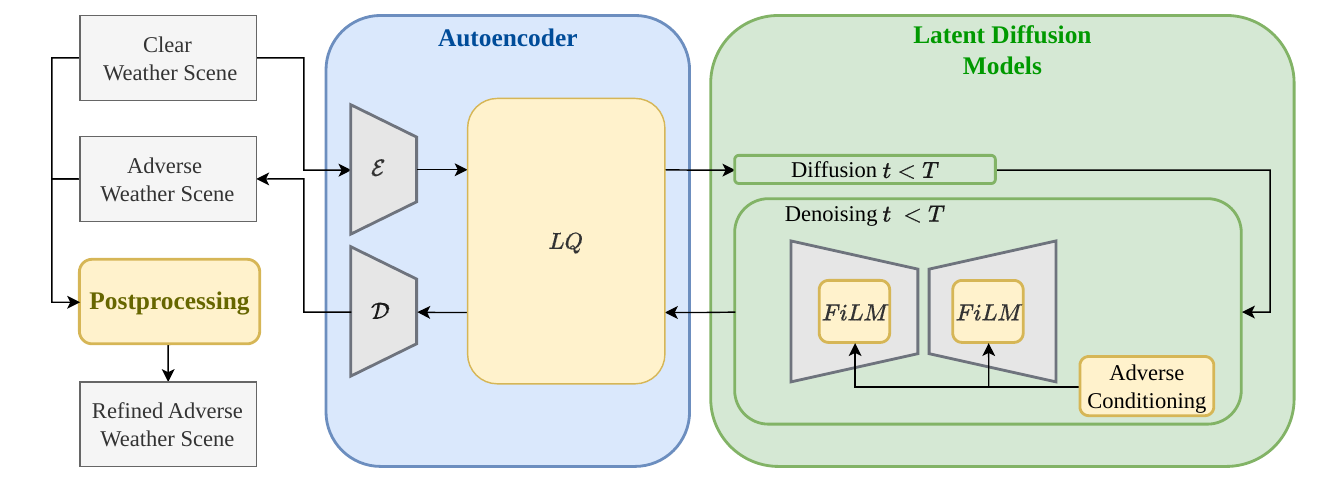}
\caption{Our method, constituted by autoencoder, latent diffusion models and postprocessing. We highlight in yellow our main contributions.}
\label{fig:01}
\end{figure}

Our method takes as input a clear weather range image of the $3$D point cloud) and generates adverse weather through a latent diffusion process. Our autoencoder generates discrete latent space with a latent quantization ($LQ$) layer. Moreover, the latent space is diffused and denoised for $t$ steps by our latent diffusion models. The denoising process is guided by adverse conditioning, which is handled by feature-wise linear modulation ($FiLM$) layers. These layers process the adverse conditioning to guide the generation of a latent space representing adverse weather while recovering the underlying structure of the clear weather scene. Finally, the adverse weather latent space is reconstructed with the autoencoder. The resulting adverse weather scene is refined with a postprocessing step, leveraging the input clear weather scene (see Fig.~\ref{fig:01}).
\subsection{Data Representation}
\begin{figure}[b]
\centering
  \begin{subfigure}{0.45\textwidth}
     \includegraphics[width=0.95\linewidth]{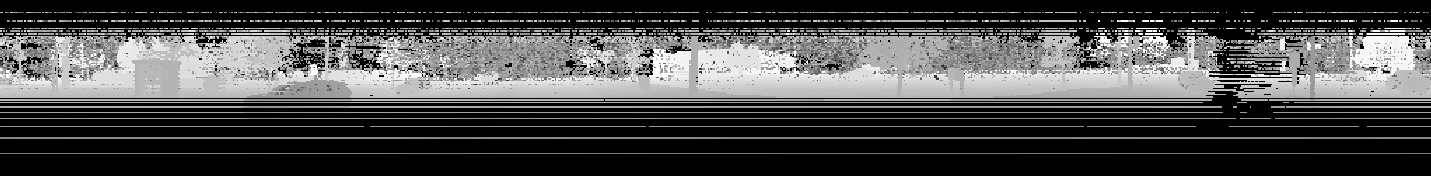}
    \caption{Uniform elevation}
  \end{subfigure}%
  \\
  \begin{subfigure}{0.45\textwidth}
    \includegraphics[width=0.95\linewidth]{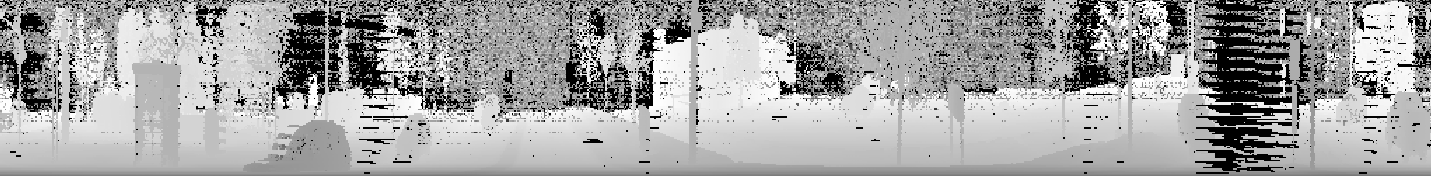}
    \caption{Beam elevation}
  \end{subfigure}%
\caption{Adverse weather depth range image. (a) with uniform elevation partition, (b) with  beam id partition (ours).}
\label{fig:02}
\end{figure}

In order to efficiently learn latent spaces via autoencoder and use diffusion models, we project point clouds in $2$D depth range images $x \in \mathbb{R}^{H \times W}$.  
The Boreas dataset \cite{burnett2023boreas} contains scenes recorded with a $128$-channel Velodyne Alpha Prime LiDAR. Differently from LiDAR sensors used in KITTI \cite{geiger2012we} and nuScenes \cite{caesar2020nuscenes}, the beams of this sensor are not equally partitioned along the elevation $\phi$. For this reason, we partition the elevation, along the image height $H$, based on the point-wise beam id $\in [0, 127]$ instead of uniformly partitioning the elevation in equal $\Delta\phi$. The azimuth $\theta$ is instead uniformly partitioned in equal $\Delta\theta$, along the image width $W$. The final projection is a one channel $128\times1024$ range image and each pixel corresponds to the normalized point depth
$d \in [-1,1]$ (see Fig. \ref{fig:02}).
\subsection{Autoencoder}
Inspired by the autoencoder proposed in \cite{ran2024towards}, we introduce an autoencoder that is trained on data from both clear and adverse weather conditions. A fundamental component of \cite{ran2024towards} is the vector quantization ($VQ$) derived from \cite{van2017neural,esser2021taming}. This component, interlayed between the encoder $\mathcal{E}$ and the decoder $\mathcal{D}$, allows to map encoded vectors of the continuous latent space (pre-quantized latent space) to a discrete space of learnable vectors in the form of a discrete spatial codebook. The mapping is achieved though a vector-wise $\argmin$ and generates a quantized latent space. Formally, given a range image $x \in \mathbb{R}^{H \times W}$, the encoder $\mathcal{E}$ performs a mapping $z = \mathcal{E}(x) \in \mathbb{R}^{h \times w \times n_z}$, where $h = H/f_h$, $w = W/f_w$, with $f_h$, $f_w$ scaling factors, and $n_z$ is the dimensionality of each spatial code $z_{ij} \in \mathbb{R}^{n_z}$.
Given a learnable discrete codebook $Z = \{z_k\}_{k=1}^{K} \subset \mathbb{R}^{n_z}$, $VQ$ performs a vector-wise mapping of each $z_{ij}$:
\begin{equation}
z_q^{VQ} = VQ(z) := \left( \argmin_{z_k \in Z} \norm{z_{ij} - z_k} \right) 
\in \mathbb{R}^{h \times w \times n_z}\;\text{.}
\end{equation}
\noindent
This discrete space helps to prevent overfitting and, compared to the continuous pre-quantized latent space, provides a simpler distribution for downstream latent diffusion models to learn from \cite{van2017neural,wang2023binary}. However, when trained on clear and adverse weather conditions, this discrete latent space does not guarantee that two similar static scenes, differing only for weather conditions, are also close in the quantized latent space. To achieve this similarity, we need to enforce a disentanglement of the codebook, in order to force the latent space to encode the adverse weather as a disentangled feature of the vectorized latent space. Following the research of \cite{hsu2024disentanglement}, we use a disentangled quantization called latent quantization ($LQ$). Thus, instead of learning a vector-wise codebook, we learn component-wise scalar codebooks. Each component of spatial codes $z_{ij}$ is mapped through a component-wise $\argmin$ over the corresponding component-wise codebook. Formally, we define each scalar codebook as a set $C_n = \{c_r \in \mathbb{R} \,|\, r=1,...,r_c\}$ of $r_c$ learnable real numbers. A spatial code vector $z_{ij}  \in \mathbb{R}^{n_z}$ can be represented as the cartesian product of $n_z$ distinct scalar codebooks $z_{ij} = \prod_{n=1}^{n_z} C_n := C_1 \times \cdot \cdot \cdot \times C_{n_z}$. Given $n_z$ learnable scalar codebooks, $LQ$ performs a component-wise mapping of each $z_{ij}$:


\begin{equation}
\begin{aligned}
&z_l = \left(\prod_{n=1}^{n_z} \argmin_{c_{r} \in C_n} \norm{z_{ijn} - c_{r}} \right)
\in \mathbb{R}^{h \times w \times n_z}\;\text{,} \\
&z_q^{LQ} = LQ(z) := z + \text{StopGradient}(z_l - z) 
\in \mathbb{R}^{h \times w \times n_z}\;\text{.}
\end{aligned}
\end{equation}
$LQ$ provides greater expressiveness by allowing combinatorial selection of the components in its codebooks. By limiting the total number of learnable scalar values in each codebook, we can enforce the autoencoder to disentangle the latent space and to assign similar components to inputs sharing corresponding similar scene features. In this way, similar scenes that differ only in weather conditions exhibit the same behavior in the quantized latent space. Forcing adverse weather into a specific vector component enables the encoding of unique features of different weather conditions and prevents the loss of high-frequency details. Moreover, forcing the latent space to map input scene features to specific vector components allows the decoder to focus independently on different scenes features \cite{hsu2024disentanglement}. Finally, disentangled adverse weather conditions allows conditioned diffusion models to associate particular distributions of the latent space to corresponding adverse weather conditions. To learn meaningful latent spaces, the autoencoder is trained on a reconstruction loss $\mathcal{L}_\textit{rec}$ between the input $x \in \mathbb{R}^{H \times W}$ and the reconstruction $\hat{x}$: 
\begin{equation}
\hat{x}=\mathcal{D}(z_q^{LQ})= \mathcal{D}(LQ(\mathcal{E}(x))\in \mathbb{R}^{H \times W}\;\text{.}
\label{eq:rec}
\end{equation}
We use only $\mathcal{L}_\textit{1}$ loss as reconstruction:
\begin{equation}
\mathcal{L}_\textit{rec}(x) = \mathbb{E}_x[\norm{x - \hat{x}}]\;\text{.}
\end{equation}
This allows to prevent overfitting, especially with adverse weather conditions reconstruction. On the other hand, relying only on $\mathcal{L}_\textit{1}$ loss limits the reconstructions of high-frequency details, since this loss focuses on low-frequency details by operating in pixel-wise distances. However, the autoencoder is trained in order to recover the structure and adverse weather conditions, while the postprocessing can restore fine-grained details. In order to learn disentangled codebooks, the autoencoder is also trained with $\mathcal{L}_\textit{quantize}$ and $\mathcal{L}_\textit{commit}$ \cite{van2017neural,esser2021taming,hsu2024disentanglement}:
\begin{equation}
\mathcal{L}_\textit{quantize} = \norm{\text{StopGradient}(z) - z_q^{LQ}}_2^2\;\text{,}
\end{equation}
\begin{equation}
\mathcal{L}_\textit{commit} = \norm{z - \text{StopGradient}(z_q^{LQ})}_2^2\;\text{.}
\end{equation}
$\mathcal{L}_\textit{quantize}$ moves the quantized vectors $z_q^{LQ}$
towards the encoder outputs $z$ and $\mathcal{L}_\textit{commit}$ makes sure the encoder commits to
a latent space \cite{van2017neural}. The total loss $\mathcal{L}$ of the autoeencoder is:
\begin{equation}
\mathcal{L} = \mathcal{L}_\textit{rec} + \mathcal{L}_\textit{quantize} +  \mathcal{L}_\textit{commit}\;\text{.}
\end{equation}
\subsection{Latent Diffusion Models}
Latent diffusion models \cite{ho2020denoising} and more recent stable diffusion \cite{rombach2022high} learn distribution of the discrete quantized latent space in order to generate latent vectors to be decoded by the pre-trained autoencoder. The generation of latent vectors can be guided towards specific conditions \cite{zhang2023adding}. Stable diffusion \cite{rombach2022high} conditions the diffusion model through cross-attention layers \cite{vaswani2017attention} along a temporal embedding of the denoising step, combined with a conditioning embedding. The use of cross-attention allows to learn strong relations between specific distributions of the latent space and the specific conditioning. In particular, the attention mechanism enables the learning of local and fine-grained dependency features. This requires a large amount of data. In adverse weather conditions, the scarcity of data, combined with the assumption that adverse weather acts as a global feature along the range image, leads us to simplify the handling of the conditioning. Starting from the architecture of stable diffusion, we replace cross-attention modules with feature-wise linear modulation ($FiLM$) layers \cite{perez2018film}. These layers are applied as intermediate modules along the U-net architecture of the diffusion model. Taking as input the time embedding $e_t$ concatenated with an adverse conditioning in the form of a binary label $b$ ($0$ for clear and $1$ for adverse), this layer learns a time-dependent conditioned linear modulation of the intermediate activations. This is achieved by learning functions $f_i$ and $h_i$ which output $\gamma_i$, $\beta_i \in \mathbb{R}^{n_i}$, that are globally applied along the intermediate activations $a_i \in \mathbb{R}^{h_i \times w_i \times n_i}$:
\begin{equation}
\begin{aligned}
\gamma_i = f_i( e_t\oplus b) \quad \beta_i = h_i( e_t\oplus b)\;\text{,}\\
FiLM(a_i | e_t\oplus b) = \gamma_i a_i + \beta_i\;\text{.}
\end{aligned}
\end{equation}
$FiLM$ allows to learn global features of adverse weather conditions, and the linearity of the layer does not require huge amount of adverse weather inputs to learn distinctive features between clear and adverse weather conditions. We train the diffusion model on both clear and adverse weather conditions scenes for a number of steps $T$ with classifier-free guidance \cite{ho2022classifier}.
The generation of latent space, derived from an iterative denoising process, is usually initialized from gaussian noise. In our augmentation method, we want to add adverse weather condition features without losing the specific scene environment. By diffusing clear weather latent spaces for a number of steps $t < T$ we can preserve feature details of the original scene and by denoising the noisy latent space for $t$ steps through the diffusion model, conditioned on snow ($b=1$), we can recover the input scene structure and add the adverse weather condition features.
\subsection{Postprocessing}
Latent spaces inherently discard high-frequency details because of their lower dimensionality space \cite{rombach2022high}, compared to the original scene $x$ $\in \mathbb{R}^{H \times W}$. During the latent diffusion process, the $t$ steps of diffusion and denoising of input latent space further degrades high-frequency details and the corresponding decoded adverse weather scene $y$ $\in \mathbb{R}^{H \times W}$ intrinsically lacks fine-grained details. Our postprocessing step leverages input clear scenes, and by distinguishing adverse features from static environment, recovers fine-grained details of the scenes while preserving the generated adverse weather conditions. Scenes generated from the latent diffusion process recover the static environment of corresponding clear scenes with the addition of adverse weather features. In the range image domain, pixels of the generated adverse weather scene representing recovered static environment are more similar to corresponding input clear scene pixels, compared to pixels representing adverse weather features. Additionally, adverse weather features decreases for the increasing sparsity of point clouds in higher depth. By deriving a depth-dependent threshold $d_t$ $\in \mathbb{R}^{H \times W}$, we can distinguish adverse weather features from the static environment at the pixel level. In particular, this threshold increases for higher depth as adverse weather features decrement, combined with the decremental precision of the latent diffusion process in recovering the structure for higher depth. By matching depths of the input clear weather scene $d_x$ $\in \mathbb{R}^{H \times W}$ and the corresponding augmented adverse weather scene $d_y$ $\in \mathbb{R}^{H \times W}$, we can take pixels representing static environment from the input clear weather scene $x$ and pixels representing occluding adverse weather features from the augmented adverse weather scene $y$. The postprocessing method is given in Algo.~\ref{alg:post}. In lines $1$-$2$, we define $d_x$ and $d_y$ as the unnormalized absolute pixel-wise depths of respectively $x$ and $y$. Moreover, in line $3$ we define $d_t$ as a pixel-wise parametrized exponential function, increasing with the depth $d_y$. A parametrized exponential function ensures low threshold values for shallow depths and a progressive increase for greater depths. In this manner, snow features are preserved at lower depths, while fine-grained details are recovered at greater depths. Finally, in line $9$, $minDepth$ selects the adverse pixel $y[i]$ only if it constitutes occlusion (lower depth than $x[i]$).
\begin{algorithm}[h!]
\DontPrintSemicolon
  
  \KwData{\\ $x \gets$ input clear weather scene \\ $y \gets$ augmented adverse weather scene \\ $\lambda, \nu \gets$ depth threshold parameters}
  \KwInput{$x$, $y$}
  \KwOutput{Refined augmented adverse weather scene $y'$}
  
  $d_x\gets Depth(x)$\\
  $d_y\gets Depth(y)$\\
  $d_t \gets exp(d_y, \nu) \cdot \lambda$\\
  $\delta = abs(d_x - d_y)$\\
  
  \For{$i \in shape(\delta)$}{
  
        \If{$ \delta[i] <  d_t[i]$}{
             $y'[i] \gets x[i]$\\

        }
        \Else{
                $y'[i] \gets minDepth(x[i], y[i])$\\
                
        }
    }
  \Return{$y'$}\;
\caption{Postprocessing}
\label{alg:post}
\end{algorithm}
\section{Experiments}
\subsection{Evaluation Metrics}
 The Chamfer distance (CD) and the Jensen-Shannon divergence (JSD) are two common metrics to evaluate reconstruction fidelity of 3D point clouds \cite{ran2024towards,nunes2024scaling}. The CD evaluates the completion at point level,
 measuring the level of detail of the reconstructed scene by calculating how far are its points from the ground truth scene.
 The JSD is a statistical metric that compares the points distribution between the reconstruction and the ground truth scene. 
Differently from previous work, we applied the JSD on the $3$D voxelized scenes without projecting to a birds-eye view (BEV). Additionally, we set a grid resolution of $0.15$m. This because we want to assess the $3$D adverse weather reconstruction quality.
\subsection{Autoencoder}
\begin{figure}[hb]
\centering
  \begin{subfigure}{0.23\textwidth}
  \setlength{\fboxrule}{0.5pt}
    \framebox{\includegraphics[width=0.9\linewidth]{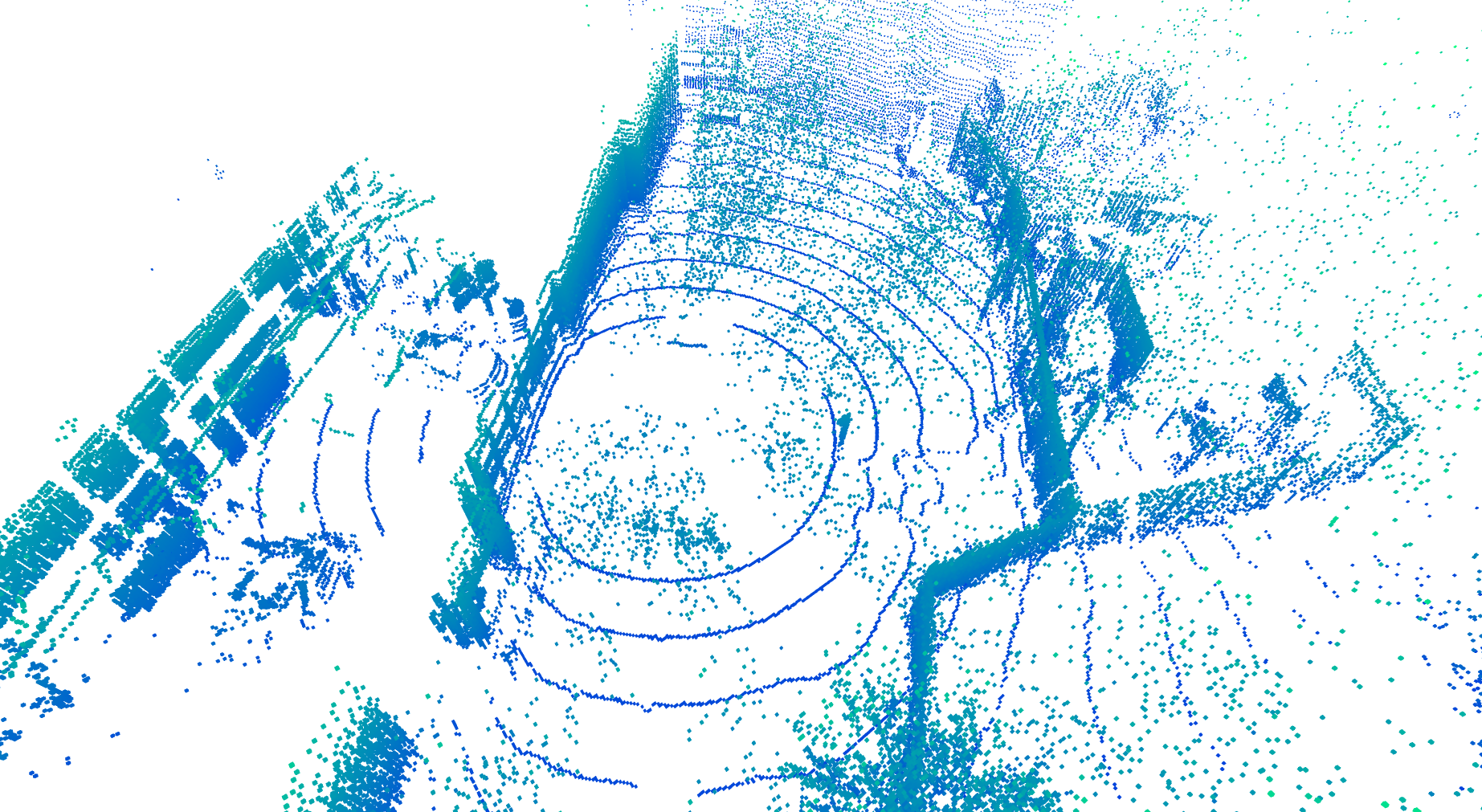}}
    \caption{Original scene}
  \end{subfigure}%
  \\[1.5ex]
  \begin{subfigure}{0.23\textwidth}
  \setlength{\fboxrule}{0.5pt}
    \framebox{\includegraphics[width=0.9\linewidth]{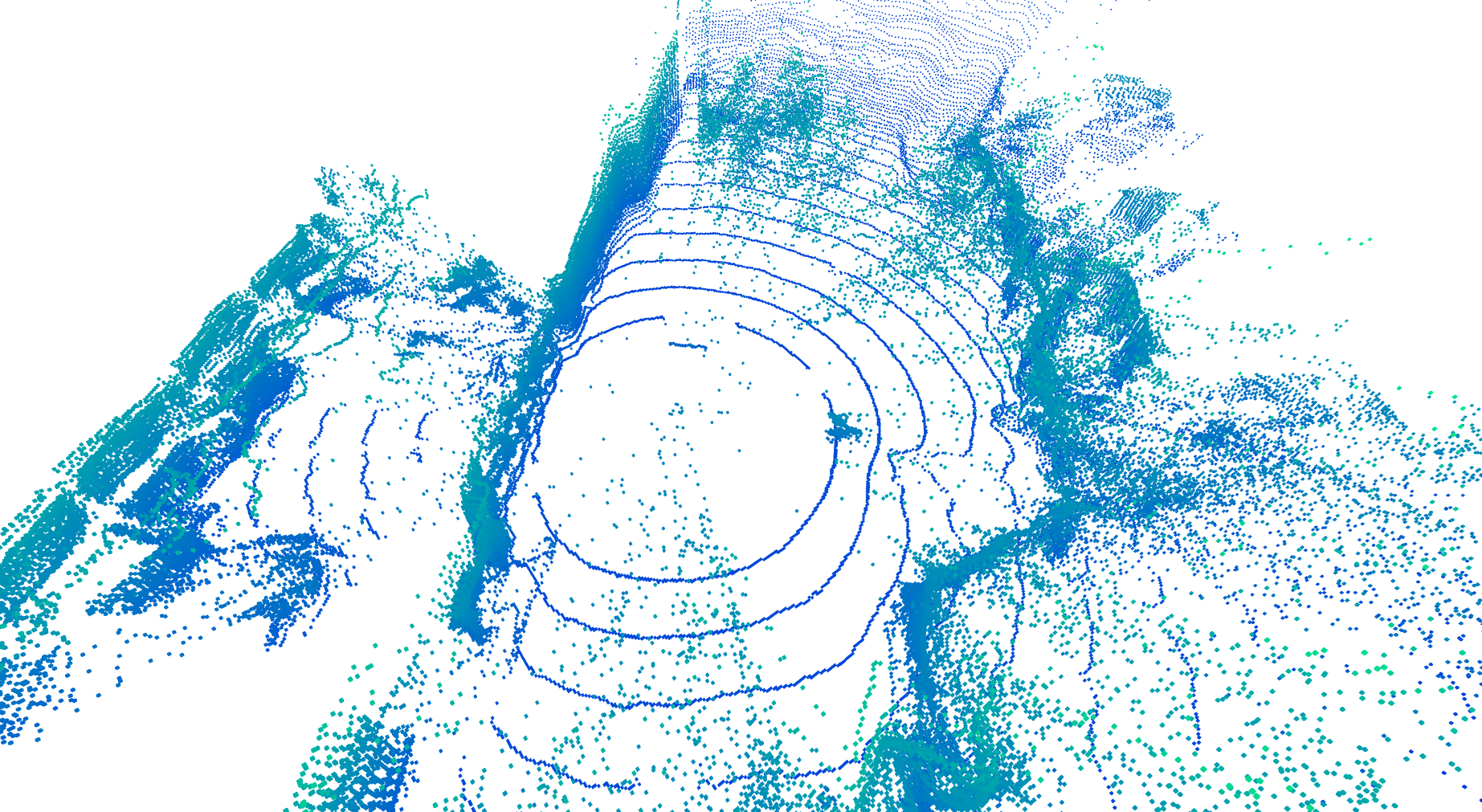}}
    \caption{$VQ$ (baseline)}
  \end{subfigure}%
  \begin{subfigure}{0.23\textwidth}
  \setlength{\fboxrule}{0.5pt}
    \framebox{\includegraphics[width=0.9\linewidth]{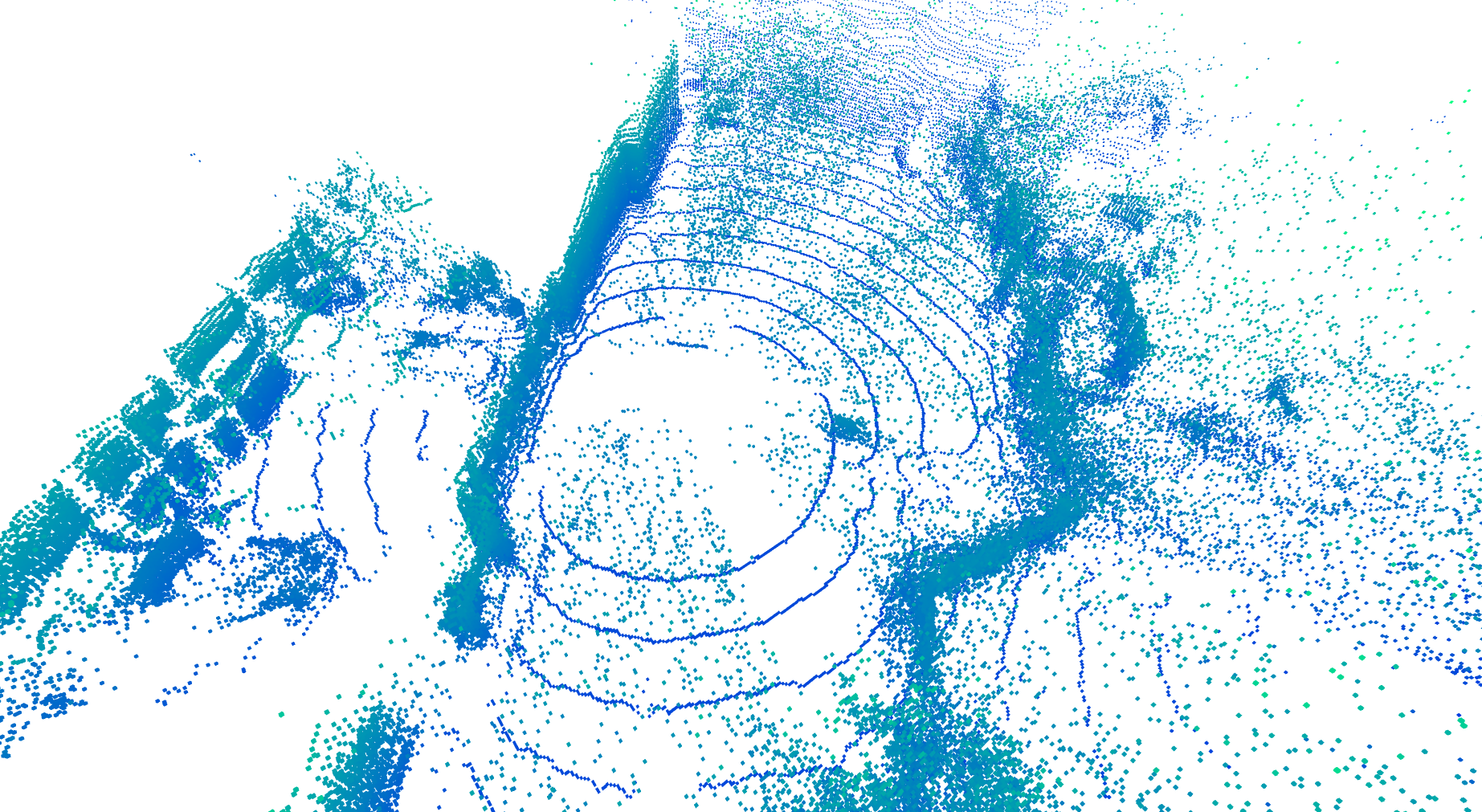}}
    \caption{$LQ$ (ours)}
  \end{subfigure}%
\caption{Snowy weather scene (a) and reconstructions (b), (c).}
\label{fig:03}
\end{figure}

We trained autoencoders with both $VQ$ and $LQ$ quantizations. We set the scaling factors $f_h=4$, $f_w=8$, and the dimensionality of each spatial code $n_z=16$ such that $z_q^{VQ}$, $z_q^{LQ}$ $\in \mathbb{R}^{32 \times 128 \times 16}$. For $VQ$, the total number of learnable vectors in the learnable discrete codebook is $\abs{Z}=16384$ and the total number of learnable values is therefore $\abs{Z}\cdot n_z=262,144$. For $LQ$, each of the $n_z$ learnable scalar codebooks learns $\abs{C_n}=256$ scalar values and the total number of learnable values is therefore $\abs{C_n}\cdot n_z = 4096$. We evaluate both autoencoders in sunny and snowy scene reconstructions. As shown in Table~\ref{autoencoder}, $LQ$ outperforms $VQ$ in the reconstruction of both sunny and snowy scenes. In both autoencoders, the reconstruction of snowy scenes is more challenging than sunny scenes. From Fig.~\ref{fig:03}, $LQ$ demonstrates higher quality of snow reconstruction. In both reconstructions, the scene structure degrades with increasing depth but the overall structure is recovered.  

\begin{table}[hbt!]
\begin{center}
\caption{Performance of autoencoders $VQ$ (baseline) and $LQ$ (ours) in sunny and snowy reconstructions.}
\label{autoencoder}
\scalebox{0.9}{
\begin{tabular}{lllllll}
\hline
&&\multicolumn{2}{c}{sun} &&\multicolumn{2}{c}{snow} \\
\cmidrule(lr){3-4} 
\cmidrule(lr){6-7}

Method && CD$\downarrow$ & JSD$\downarrow$ && CD$\downarrow$ & JSD$\downarrow$ 
\\ 
\hline & \\[-1.5ex]
 $VQ$ (baseline) & & $0.276$ & $0.691$ & & $0.281$ & $0.692$\\
 $LQ$ (ours) & & $0.181$ & $0.647$ & & $0.194$ & $0.659$
\end{tabular}
}
\end{center}
\end{table}

\bibliography{mybibliography}
\bibliographystyle{unsrt}

\end{document}